\documentclass{article}

\usepackage[preprint]{corl_2024} % Use this for the initial submission.
\usepackage{amsmath}
\usepackage{bm}
\usepackage{capt-of}
\usepackage{amssymb}
\usepackage{mathtools}

\usepackage{cuted}
\usepackage{booktabs}
\usepackage{hyperref}

\usepackage[normalem]{ulem}

\newcommand{\ie}{\textit{i.e.}}
\newcommand{\eg}{\textit{e.g.}}
\newcommand{\basep}{$\bm{\pi}_b$ }
\newcommand{\taskp}{$\bm{\pi}_t$ }
\newcommand\mypara[1]{\vspace{1mm}\noindent\textbf{#1}}

\newcommand{\revision}[1]{{\color{black}{#1}}}
\newcommand{\secrevision}[1]{{\color{black}{#1}}}
\usepackage{xcolor}
\title{Hand-Object Interaction Pretraining from Videos}

\author{Himanshu Gaurav Singh$^*$\And Antonio Loquercio\thanks{denotes equal contribution. All authors are affiliated to UC Berkeley.}%
\And Carmelo Sferrazza\And Jane Wu\And Haozhi Qi\And Pieter Abbeel\And Jitendra Malik} % <-this % stops a space

\begin{document}
\maketitle

% From HHOT to DRP
%Hand object interaction pretraining from videos
%===============================================================================

\begin{abstract}
We present an approach to learn \textit{general robot manipulation priors} from 3D hand-object interaction trajectories. We build a framework to use in-the-wild videos to generate sensorimotor robot trajectories. We do so by lifting both the human hand and the manipulated object in a shared 3D space and retargeting human motions to robot actions. Generative modeling on this data gives us a task-agnostic base policy. This policy captures a general yet flexible manipulation prior. We empirically demonstrate that finetuning this policy, with both reinforcement learning (RL) and behavior cloning (BC), enables sample-efficient adaptation to downstream tasks and simultaneously improves robustness and generalizability compared to prior approaches. Qualitative experiments are available at: \hyperlink{https://hgaurav2k.github.io/hop/}{https://hgaurav2k.github.io/hop/}.
\end{abstract}


%===============================================================================
\section{Introduction}

Reusable sensorimotor representations have the potential to give robots access to the versatility of their sensorimotor apparatus, thereby enabling them to achieve a wide variety of goals.
Similar to advancements in other AI domains~\cite{he2022masked,radford2019language}, such representations are likely to be trained with unsupervised objectives on large datasets.
In this work, we study the feasibility of training such representations using human videos in the context of dexterous manipulation.

Using videos as a data engine comes with several advantages: (1) they are abundant; (2) they cover a wide range of skills that we want robots to master; and (3) they reflect natural or socially acceptable behaviors that we want robots to emulate.
However, training sensorimotor representations on videos is a challenging endeavor.
First, videos only partially capture the nature of an agent's interaction with their surroundings.
For instance, by looking at a person holding an object, it is almost impossible to estimate the force their fingers are exerting.
In addition, the larger the embodiment gap between a human and a robot, the more their actions will differ to achieve the same objectives.

The difficulty of learning from videos led previous work
to mostly focus on specific aspects of the problem. One line of research focused on training visual representations with off-the-shelf self-supervised vision algorithms on large vision datasets~\cite{ma2022vip,Radosavovic2022,majumdar2024we,dasari2023unbiased, nair2022r3m}. While simple and effective, such pretrained representations lack a motor component, making them less effective on downstream tasks~\cite{radosavovic2023robot}. Another line of work aims to extract both sensory and motor information from videos by estimating human motions in 3D~\cite{shaw2023videodex, peng2018sfv, patel2022learning, qin2022dexmv, radosavovic2024humanoid}. However, these approaches require alignment between the training videos and the robot's downstream tasks, which compromises the generality of the learned representations. \secrevision{Finally, recent works aim to use egocentric videos of human activities to learn an explicit hand-object interaction prior in the form of a contact-pose prediction model~\cite{kannan2023deft,mandikal2022dexvip}. While a contact-pose prior is potentially task-agnostic, useful information in hand-object trajectories extends beyond contact-poses, including but not limited to pre/post-contact trajectories, intuitive physics of the interaction and human preferences.}

\begin{figure}[t]
    \centering
    \includegraphics[width=\textwidth]{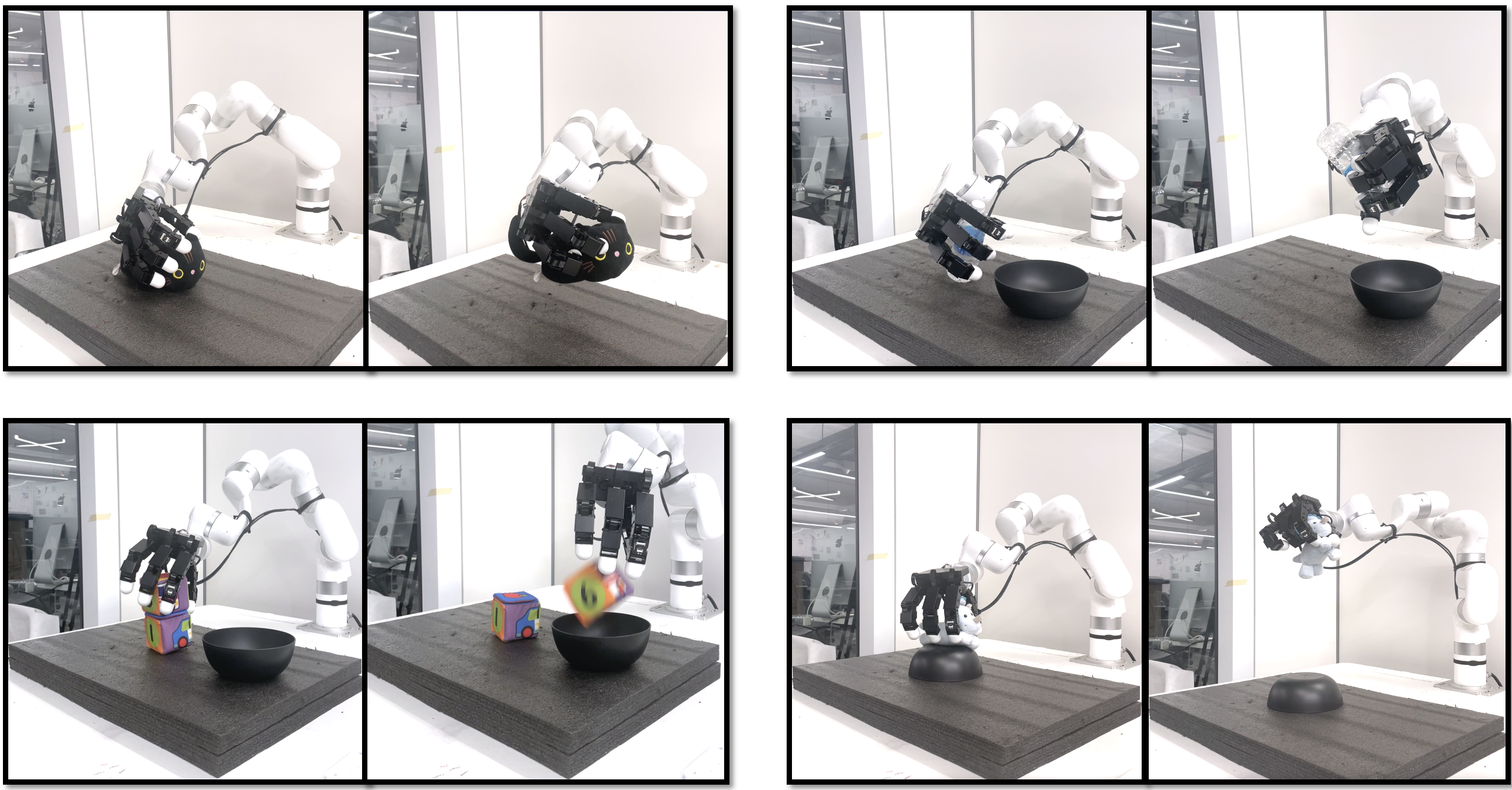}
    \caption{\footnotesize{Real world rollouts of the policy finetuned from HOP using less than 50 demonstrations. HOP enables sample-efficient downstream adaptaion by learning a general manipulation prior from human videos.}}
    \label{fig:figure1}
\end{figure}

In this paper, we present an approach to capture a general manipulation prior from in-the-wild videos. Such a prior is implicitly embedded in the weights of a causal transformer, pretrained with a conditional distribution matching objective on sensorimotor robot trajectories. These trajectories are generated by mapping 3D hand-object interactions to the robot's embodiment via a physically grounded simulator. \secrevision{The choice of an implicit prior, aligned with the current paradigm in vision and language research, has the potential advantage of becoming more and more expressive as the quality and diversity of the data increases.} The resulting prior can be quickly adapted to any manipulation task either with reinforcement learning or behavioral cloning. After adaptation, the prior takes the form of an \emph{end-to-end} policy mapping the robot's multi-modal sensory stream to low-level joint commands. This policy can be directly executed on a physical robot (see Fig.~\ref{fig:figure1}).

We empirically study the advantages brought forward by pretraining with hand-object interactions in both simulation and real-world experiments. The findings of this study indicate that our manipulation prior considerably speeds up skill acquisition compared to previous methods, even if such skills are not represented in the training videos. Additionally, it improves generalization and robustness to disturbances in the downstream policy.
% These findings suggest a promising path toward learning
% challenging real-world robot control tasks by generative
% modeling of large collections of sensorimotor trajectories.

\section{Overview}
\label{sec:overview}
% \method uses videos of people interacting with objects to train a base policy for a dexterous robot.

The objective of \textbf{H}and-\textbf{O}bject interaction \textbf{P}retraing (HOP) is to capture general hand-object interaction priors from videos.
% Such base policy is used as a foundation for downstream tasks, on which it is fine-tuned either with reinforcement learning or behavior cloning.
In contrast to previous work, we do not assume a strict alignment of the human's intent in the video and the downstream robot tasks.
Our key intuition is that the basic skills required for manipulation lie on a manifold whose axes are well covered by unstructured human-object interactions.

We extract sensorimotor information from videos by lifting the human hand and the manipulated object in a shared 3D space. We then bring such 3D representations to a physics simulator, where we map human motion to robot actions.
There are several advantages to using a simulator as an intermediary between videos and robot sensorimotor trajectories: (i) we can add physics, inevitably lost in videos, back to the interactions; (ii) it enables the synthesis of large training datasets without putting the physical platform in danger; and (iii) we can add diversity to the data by randomizing the simulation environment, e.g., varying the friction between the robot's joints, the scene's layout, and the object's location relative to the robot.

We generate a dataset of robot-object interactions $\mathcal{D}=\{\tau_1, \tau_2, \ldots, \tau_N\}$ where 
$\tau=\{ (\bm{o}[0], \bm{a}[0]), (\bm{o}[1], \bm{a}[1]), \ldots, (\bm{o}[T], \bm{a}[T]) \}$ are the observation-action pairs of a single sensorimotor trajectory.
An observation $\bm{o}[k] \in \mathbb{O}$ at time $k \in [0,\dots, T]$ consists of \revision{visual scene information (a depth image or a pointcloud)} and robot's joint angles $\phi[k]$, \ie, its proprioception.
The action $\bm{a}[k] \in \mathbb{A}$ consists of continuous joint angles, which are converted to joint torques with a low-level PD controller. 
We use $\mathcal{D}$ to train a base policy $\bm{\pi}_b$ on the unsupervised objective of next-action prediction from a history of sensory observations, \ie, $\hat{\bm{a}}=\bm{\pi}_b(\bm{o}[t:t-L])$, where $L$ is a fixed context length.
%
% We instantiate $\bm{\pi}_b$ as a causal transformer architecture~\.
%
% When deployed on a real robot, \basep exhibits primitive manipulation abilities, \eg, moving the end-effector towards an object and using its fingers to interact with it.
We finetune \basep to generate task-specific policies \taskp either optimizing a reward with reinforcement learning or a behavioral cloning objective on few task-specific demonstrations. The next section presents each aspect of our method in detail.

% We propose \textbf{H}uman hand-object \textbf{T}rajectory \textbf{P}retraining. As described in figure \red{add figure}, we extract 3-D hand-object trajectories from human manipulation videos and retarget it to a robot embodiment. We train a transformer-based policy to predict robot actions given past-history of joint states and object point clouds using this data. This policy learns general hand-object interaction priors. \red{On zero-shot deployment, it can reach objects with human-like affordance}. Finetuning this policy on few-shot demonstrations enables success on downstream tasks.  
% \red{the assumptions on the dataset, what data is finally used. There is a single object, single hand...manageable camera motion. Need to formalise} We finally use a dataset of ~600 videos consisting of the DexYCB dataset \cite{Chao2021DexYCBAB} and a subset of 100 videos from the MOW \cite{rhoi2020} dataset. 

\section{Method} 
\begin{figure}[t]
    \centering
    \includegraphics[width=\textwidth]{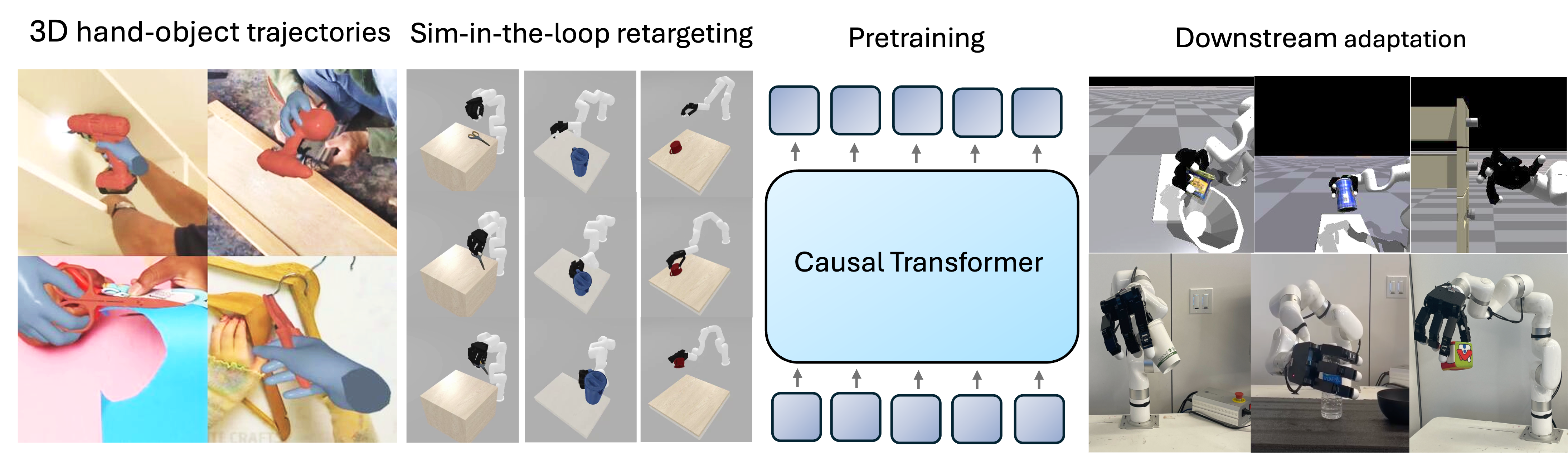}
    \caption{\footnotesize{3-D hand-object trajectories from in-the-wild human manipulation videos are re-targeted to a robot embodiment within a physics simulator, resulting in physically grounded robot data. General manipulation priors are learnt from this using generative modelling of trajectories. Such representation enables sample-efficient adaptation for new downstream tasks.}}
    \label{fig:approach}
\end{figure}

\subsection{Lifting Hand-Object Interaction Videos to 3D}
\label{sec:pose-estimation}

Recovering the underlying 3D structure of hand-object interactions from in-the-wild monocular videos is inherently ambiguous.
To alleviate such ambiguity, previous work leveraged the insight that the human hand can be used as an anchor for the 3D location and scale of the manipulated object~\cite{wu2024reconstructing,choi2023handnerf,ye2022s, tekin2019h+,hasson2019learning,cao2021reconstructing}.
Our setup to estimate hand-object interaction trajectories from videos builds upon recent advances in 3D vision. Our approach closely follows MCC-HO~\cite{wu2024reconstructing} with a few modifications to adapt it to our use case.

Given a single RGB image and an estimate of the 3D hand geometry from HaMeR~\cite{pavlakos2024reconstructing}, MCC-HO jointly infers hand-object geometry as point clouds. To fine-tune the quality of the prediction, MCC-HO finetunes the object's pose by fitting it to a CAD model. However, this finetuning assumes knowledge of the object the human is interacting with. To increase generality, we wave this assumption and skip the CAD-based post-processing.
This simplification comes at the cost of reduced reconstruction quality and temporal smoothness.
While we find the first problem not critical for pre-training, we increase temporal smoothness by anchoring object reconstructions to time-smoothed hand detections~\cite{pavlakos2024reconstructing}. In addition, we make the simplifying assumption that the camera from which the video is collected is static. More details of our 3D estimation pipeline are provided in the appendix. The result of this pipeline is a sequence of 3D hand-object poses.

\subsection{Mapping 3D Human-Object Interactions to Robot-Object Interactions}
\label{sec:retargeting}

We formulate a non-linear optimization problem to generate a sensorimotor trajectory $\tau$ from a sequence of 3D hand-object poses. At each step $k$, we find the action $\bm{a}[k]$ by optimizing the following cost function:
\begin{equation}
    \min_{\bm{a}[k]} \frac{1}{2}\| \bm{x}_{h}[k] - f(\bm{a}[k]) \|^2 + \lambda \| \bm{a}[k] - \bm{\phi}[k-1] \|^2 \quad \text{s.t.} \quad \bm{a}[k] \in \mathbb{A},
    \label{eq:optim}
\end{equation}
where $f$ is the robot's forward kinematics, and $\bm{x}_{h}[k]$ are the 3D coordinates of a set of keypoints on the human hand.
% (shown in Fig.~\ref{img:retargeting}) \red{add re-targeting image}.
\revision{The first term of Eq.~\ref{eq:optim} represents the difference between the robot’s and the human keypoints as a function of the robot’s desired joints $\bm{a}[k]$. The second term is proportional to the energy required to execute the action $\bm{a}[k]$, which we minimize to favor smoothness.}

While there are approaches that include the object dynamics in the optimization~\cite{zhu2023diff,lakshmipathy2023contact,lakshmipathy2024kinematic,kim2016retargeting}, they are challenging to apply to in-the-wild videos due to noise in object pose estimates. \revision{In addition, in-the-wild videos do not have reliable information about objects' physical properties, \eg, mass or friction.} Therefore we disregard the dynamics of the manipulated object and place it on every step at the location observed in the video. While this can lead to physical implausibility in the object motion and possibly lack of force-closure grasps, the data quality does not deteriorate much in practice, as we can still learn useful behaviors. \revision{We empirically show that without object-trajectories (Section \ref{sec:hand-only}), the quality of the base policy decreases.}

The optimization is performed independently on each timestep $k$, and the resulting actions $\bm{a}[k]$ are executed in a high-fidelity simulator to generate $\bm{o}[k]$. We refer to this method of mapping human motion to robot sensorimotor trajectories $\tau$ as \textit{simulator-in-the-loop retargeting}. The primary advantage of this approach is the optimization in~\eqref{eq:optim} can be conducted using a simplified forward kinematics $f$, reducing the computational burden. Despite this simplification, the actions are executed in a high-fidelity simulator, ensuring realistic behaviors and high-quality observations.

We randomize the simulated scene to increase data diversity. Specifically, we add obstacles like tables and walls to the scene and vary their positions relative to the robot. \revision{This allows us to add random constraints to this optimization problem, which increases the overall diversity in the extracted joint trajectories. This is particularly important for robots with kinematic redundancies, since they have multiple joint position trajectories for the same end-effector trajectory.}
Note that this approach to retargeting differs from previous methods, disregarding other objects in the scene and optimizing actions via physics-based constraints, \eg, minimum jerk~\cite{qin2022dexmv,sivakumar2022robotic,shaw2023videodex} or minimum velocity~\cite{radosavovic2024humanoid}. 

The quality of the resulting robot trajectories decreases as the difference between the environment where the video was collected, the simulated scene, and $f$ grows.
However, given the non-convex optimization landscape of ~\eqref{eq:optim}, we can obtain good trajectories by running the optimization multiple times with various initial positions and scene layouts. High-quality data is then obtained by pruning the trajectories on metrics like collision with obstacles and the tracking error between the hand's and the robot's keypoints.

% \begin{itemize}
%     \item Describe the retargeting procedure
%     \item Describe the details of object generation
%     \item Describe how trajectories are pruned
% \end{itemize}

% \red{following lines can be moved to appendix}
% Broadly, we translate the humand hand trajectories to robot joint trajectories such that the robot fingertips mimic the trajectory of those of the human. \citet{Qin2021DexMVIL} uses backpropagation across the forward dynamics model to optimize for the joint trajectories of the robot, with constraints being the end-effector pose (in our case, the fingertips' pose). The forward dynamics model consists of gravitational forces on the hand and object and the internal forces between robot links. Note that object and hand contact dynamics are non-differentiable and thus hard to optimize, therefore, we ignore these during the retargeting optimization. While this can lead to physical implausibility in the final trajectories, in practice we find that this does not deteriorate the data quality as much and we can still learn useful behaviors. Access to 3-D poses of the hand and object allows us to  augment these trajectories by randomizing the starting pose of the hand and object . We generate 40,000 hand-object trajectories from 600 videos. 

% Finally, we have a dataset of trajectories $\mathcal{T}=\{\tau_1, \tau_2, \ldots, \tau_N\}$ where 
% $\tau=\{ (j_0, p_0, a_0), (j_1, p_1, a_1), \ldots, (j_T, p_T, a_T) \}$. Here $j_t,p_t,a_t$ is the robot proprioception, object point cloud and robot action respectively. 

\subsection{Robot Trajectory Pretraining}
\label{sec:architecture}

The resulting trajectory dataset $\mathcal{T}$ contains knowledge that could be valuable to any manipulation tasks. For instance, $\mathcal{T}$ has information about object affordance, \ie, where and how to grasp; some intuitive (although rudimentary) physics, \eg, an object should be reached upon before being lifted; or wrist-hand coordination, \ie, the behavior of orienting and shaping the hand simultaneously while moving the wrist to maximize efficiency~\cite{jones2006human}.

We aim to incorporate this knowledge as useful behavioral priors into a policy $\pi_b$ that can be finetuned to downstream tasks.
Similar to previous work in language~\cite{radford2019language}, vision~\cite{chen2020generative}, and robotics~\cite{radosavovic2023robot, kumar2022pre, team2024octo}, we instantiate $\pi_b$ as a transformer~\cite{vaswani2023attention} and train it on a generative modeling objective.
Specifically, we train $\pi_b$ to capture the conditional distribution $\Pi(\bm{a}[t-L:t] |\bm{o}[t-L:t])$ by optimizing the following loss:
\begin{equation}
        \mathcal{L}(\tau ; \theta) = \mathbb{E}_{t \sim [1 \ldots T]} \left[ \left\| \bm{a}[t -L:t] - \pi_b(\bm{o}[t -L:t]) \right\|_1 \right].
        \label{eq:pretrain_loss}
\end{equation}
However, unlike previous work, our pretraining dataset $\mathcal{T}$ contains neither real-world demonstrations nor complete task executions. This is because our data is generated from unstructured 3D hand-object interactions, and we disregard the dynamics of the manipulated object during retargeting (Sec.~\ref{sec:retargeting}). Yet, we find that the pre-training paradigm in~\eqref{eq:pretrain_loss} leads to the emergence of useful representations in $\pi_b$.

% \subsection{Downstream Finetuning}
% \label{sec:finetuning}

\mypara{Downstream Finetuning.}
The pretrained policy $\pi_b$ exhibits primitive manipulation skills, \eg, reaching an object with a reasonable grasp pose, while occasionally grasping successfully.
We finetune these skills to a task by optimizing a reward with reinforcement learning or a behavior cloning loss on limited demonstrations.
We finetune the whole model for the task.
Empirically, we find that finetuned policies use the information in $\pi_b$ to train faster, are more robust to disturbances, and generalize better than policies trained from scratch and a set of baselines.
In addition, we find that the finetuning process re-utilizes the information in $\pi_b$ even for tasks not explicitly represented in the training videos.

% \begin{itemize}
%     \item Describe the architecture.
%     \item Describe the tokenization.
%     % \item Describe the optimization process.
% \end{itemize}

\section{Experimental Setup}
\label{sec:exp_setup}

\mypara{Robot.} We use a low-cost 7-DoF xArm robot with a 16-DoF Allegro hand~\cite{allegro} vertically mounted at its end effector. The proprioception observation $\bm{\phi}_k$ includes joint position from both robots. While we don't make any specific assumption about the robot embodiment, we use a multi-fingered hand instead of a parallel joint gripper since demonstration quality increases as the embodiment gap between the robot and the human decreases. We empirically found that since the robot base is fixed, a 7-DoF arm can track much better human motions than a 6-DoF arm, which often encounters singularities during such trajectories. Visual sensing comes from a single stereo camera (Zed-2) mounted on the robot's right side.

\mypara{Simulation Setup.} Our simulation environment is developed with the IsaacGym ~\cite{makoviychuk2021isaac} simulator. The robot morphology and action space are identical to the real setup. However, since rendering depth images is prohibitively expensive, we give the agent access to the ground-truth object pointcould instead of a depth image (see Section ~\ref{sec:overview}). Specific details about the task setup and reward design can be found in the appendix.

\mypara{Video Datasets.} 
% \red{TODO: @Jane, can you say a few words about where our data comes from and the right references?}.
Our pretraining dataset of 3D hand-object trajectories consists of sequences from two datasets: DexYCB~\cite{Chao2021DexYCBAB} and 100 Days of Hands~\cite{shan2020understanding}. 
We use 250 videos from the DeXYCB dataset (right-hand only) annotated with ground truth hand-object trajectories as a source of high-quality data.
%
% The DexYCB dataset contains 1,000 hand-object grasping trajectories captured by eight RGB-D cameras. 
% Since the 3D hand and YCB object poses are provided, we use this ground truth data directly.
% The 100 Days of Hands dataset~\cite{shan2020understanding} is a collection of Internet videos where hands are interacting with objects.
We additionally use approximately 200 videos from the 100 Days of Hands dataset. Sixty percent of these videos were previously annotated with hand-object interaction trajectories~\cite{patel2022learning}, which we directly use.
% The vast majority of the video frames are unlabeled, but we use all the estimated trajectories provided by RHOV~\cite{patel2022learning}.
We annotate the remaining videos with our 3D estimation pipeline (Sec.~\ref{sec:pose-estimation}).
%
% HO~\cite{wu2024reconstructing} and HaMeR~\cite{pavlakos2024reconstructing} to infer 3D hand and object geometry for an additional $\sim$50 video sequences in 100DOH.
Overall, our combined dataset contains approximately 450 videos.
We retarget these videos to obtain a pretraining dataset $\mathcal{T}$ of approximately $70,000$ trajectories.

\mypara{Retargeting.} We use low-storage BFGS~\cite{liu1989limited} from the NLOpt library~\cite{NLopt} for optimization. We perform simulation-in-the-loop retargeting in a simple simulated scene with a ground floor on which the robot and a static table are placed 65cm apart. Objects start their trajectories above the table with a random pose. We run the optimization $700$ times for each video, randomizing the table location and the robot's initial joint state. We add a trajectory to $\mathcal{T}$ only if, at any time, their retargeting error (See Eq.~\eqref{eq:optim}) is below 3cm and the arm does not collide with the table or the floor. 
% We use Sapien~\cite{Xiang_2020_SAPIEN} as our simulated engine.
Our code is built upon the implementation of Qin et al.~\cite{qin2022from}.

\mypara{Transformer.} Similar to previous work~\cite{chen2021decision}, we represent the policy $\pi_b$ with a GPT-2-style causal transformer. The policy takes proprioception and observation input from the past 16 timesteps and predicts the next action. Details about the architecture can be found in the appendix.

% Specifically, the transformer has 4 layers, 4 heads, 192 embedding dimensions, and no dropout. The policy takes proprioception and observation input from the past 16 timesteps and predicts the next action. Proprioception is encoded into the transformer's hidden dimension using a linear projection and the depth observation is encoded using a CNN. Finalproject

\mypara{Pretraining.} We train the transformer with the objective in Eq.~\eqref{eq:pretrain_loss} on $\mathcal{T}$. While we could make the prediction autoregressive and add decoding heads and proxy losses for future proprioception and images (as in~\cite{radosavovic2024humanoid}), we empirically found these changes to be not very helpful in practice to our tasks. Therefore, we predict only future actions for simplicity. We use as optimizer AdamW~\cite{loshchilov2017decoupled} with initial learning rate of $10^{-4}$ and weight decay of $10^{-2}$. \revision{We trained two distinct base policies—one with depth observations and the other with point cloud observations. The former is used for real-world, and the latter for simulation experiments. However, it's important to note that both policies were trained on exactly the same trajectories; only the associated sensor observations differed.
}

\mypara{Finetuning.} \textit{In simulation}, we finetune the transformer with PPO~\cite{schulman2017proximal} using the default hyperparameters from~\cite{petrenko2023dexpbt}. However, we add a few modifications inspired by~\cite{ramrakhya2023pirlnav} for effective fine-tuning: (1) we use a small initial exploration noise of 0.1; (2) the value and policy networks share the observation tokenizer, but the tokenizer's weights are not updated by the value function's gradients; (3) we warm up the value function's parameters for the first 200 gradient steps, keeping the actor parameters fixed. \revision{Since reinforcement learning requires up to 1 billion steps to convergence, and our simulator does not offer fast multi-gpu rendering, we use pointclouds as visual information, as they can be efficiently simulated.} \textit{In the real world}, we finetune the entire $\pi_b$ on limited demonstrations with the same objective and hyperparameters used for pretraining.  \revision{In these experiments, we use depth images as input to our policy since pointcloud estimation in the real world generally requires multiple cameras, while our real-world setup has a single camera.}
% \red{any difference at BC time?} 
% \red{add more RL finetuning details like object set, environment details etc}

\mypara{Inference.} At test time, the model operates in \textit{closed-loop}: it receives the past and current observations as input and predicts the next action to execute. 
% In contrast to the predominant wisdom in behavioral cloning~\cite{loquercio2021learning, chi2023diffusion, zhao2023learning,radosavovic2023robot}, we do not predict multiple actions in the future. This enables using the same inference mechanism for reinforcement learning and behavioural cloning tasks.
The prediction loop runs at $20$Hz. The predicted action is sent to the xArm and Allegro low-level controllers, which operate at 120Hz and 300Hz, respectively.

% \mypara{Tasks} We finetune $\pi_b$ on three tasks of increasing difficulty in simulation: \texttt{Cabinet}, \texttt{Grasp and Lift}, and \texttt{Throw}. In the real world, we have three similar tasks: \texttt{Grasp and Lift}, \texttt{Throw}, and \red{\texttt{Hand-Shake}}.

\section{Experimental Results}

% Our experiments investigate the effectiveness of HOP as a base model that can be finetuned to downstream tasks. Specifically,

We design an experimental procedure to analyze the advantages brought forward by HOP in terms of finetuning efficiency,
generalization, and robustness to perturbations.
Specifically, we ask the following questions:
\secrevision{(i) \emph{How does HOP compare to vision-only pre-training approaches for robot learning?}
(ii) \emph{How does HOP compare to existing demonstration-guided reinforcement learning algorithms~\cite{rajeswaran2018learning,peng2021amp}?} 
(iii) \emph{How does learning from hand-object interaction trajectories compare to learning hand-pose priors only}~\cite{bharadhwaj2023zero,ze2024h,shaw2023videodex}}?.
We answer these questions via controlled experiments in simulation and the physical world.

\subsection{Comparison to visual pre-training baselines (real-world).} 

\mypara{Baselines.}
We compare our approach to visual pre-training systems. Such systems are trained on large image or video datasets but lack a motor component. Specifically, we compare to methods using the following pre-training data:
\begin{itemize}
    \item \textbf{ImageNet} We encode the depth image with a VIT-B network~\cite{dosovitskiy2020image} pre-trained on ImageNet and pass the resulting CLS token embeddings to our transformer. The latter is then trained with real-world data. We consider two variants: using the VIT features zero-shot (\emph{Imagenet ZS}) and finetuning them on the downstream dataset (\emph{ImageNet F}).
    \item \textbf{Internet Videos} We use off-the-shelf visual features from R3M~\cite{nair2022r3m}, VIP~\cite{ma2022vip}, and MVP~\cite{radosavovic2023robot}. These features were obtained with unsupervised contrastive learning objectives on large video datasets, e.g. Ego4D~\cite{grauman2022ego4d}. Conversely to ours, these baselines don't use depth but RGB images as input.
\end{itemize}
We additionally compare to Diffusion Policies~\cite{zhu2023diff} using a UNet backbone since our tasks exhibit temporally smooth desired action sequences. Similarly to ours, this baseline uses depth as input.
With the above baselines, we want to understand how classic methods for behavior cloning work in our setting, where a single camera and a limited number of demonstrations are available. 
Indeed, the previous approaches are generally applied with a large number of demonstrations and multiple RGB cameras.

\mypara{Tasks} We evaluate our approach in the real world on three tasks of increasing complexity. In the first task, \textit{Grasp and Drop}, the robot needs to unstack a cube and put it in a bowl.
The second is the \textit{Grasp and Pour} task, where the robot needs to pick a bottle and point it towards a bowl. In the third task, \textit{Grasp and Lift}, the robot must pick up one of 4 different-looking objects, all requiring different spatial affordances. One single model is trained to pick up all objects. In this task, we evaluate the ability of the approach to adapt with a few demonstrations on very different object shapes.
We collect 15 demonstrations for the first two tasks and 50 demonstrations for the third task. We encourage the reader to check our project page for visualizing the tasks. More details about the scene setup and evaluation criteria can be found in the appendix.

\mypara{Results} Table~\ref{tab:rw_results} summarizes the result of our study.
The findings indicate that for tasks with a single object (Grasp \& Drop, Grasp \& Pour), all methods perform comparably and achieve, with some sporadic exceptions, a close-to-perfect success rate.  
However, in the hardest task (Grasp and Lift), where a single policy needs to pick four objects with different affordances, our approach has a margin of 30 percentage points to ImageNet Finetuned, the best-performing baseline.
We additionally find that the baselines using RGB data are much more successful with a single object than with multiple ones. This is likely because there are not enough demonstrations to learn object-specific affordances.
Overall, these results empirically validate the value of our sensorimotor pretraining strategy.

\begin{table}[ht]
\centering
\begin{tabular}{l | cccc | ccc}
\toprule
 & \multicolumn{4}{c}{\textbf{Depth-input}} & \multicolumn{3}{c}{\textbf{RGB-input}} \\
 \textbf{Task} & Ours & Diffusion-Policy & Imagenet-ZS & Imagenet-F & R3M & VIP & MVP \\ 
\midrule
Grasp \& Drop  & 0.80 & \textbf{0.90} & 0.90 & 0.80 & \textcolor{gray}{0.0} & \textcolor{gray}{0.40} & \textcolor{gray}{0.20} \\
Grasp \& Pour  & \textbf{1.0} & 0.20 & 0.80 & 0.70 & \textcolor{gray}{1.0} & \textcolor{gray}{1.0} & \textcolor{gray}{1.0} \\
Grasp \& Lift  & \textbf{0.65} & 0.30 & 0.30 & 0.35 & \textcolor{gray}{0.0} & \textcolor{gray}{0.0} & \textcolor{gray}{0.0} \\
\bottomrule
\end{tabular}
\vspace{1ex}
\caption{Real-robot results (success rate \% averaged over 20 rollouts). Note that RGB baselines are not directly comparable to our approach since our policy takes depth images as input.}
\label{tab:rw_results}
\end{table}

\subsection{Comparison to demonstration-guided reinforcement learning strategies (simulation)}

Our simulation experiments investigate the effectiveness of HOP as a base model for adaptation to downstream tasks using RL. The simulation agent is identical in morphology to the real robot.

% \red{What questions would we like to answer?}
% We finetune $\pi_b$ on three tasks of increasing difficulty in simulation: \texttt{Cabinet}, \texttt{Grasp and Lift}, and \texttt{Throw}.

\mypara{Baselines.}
% We compare HTP with PPO from scratch and approaches to incorporate offline reward-free demonstrations in RL\cite{peng2021amp, rajeswaran2018learning}.
We compare our approach to three baselines: (1) training from scratch (\emph{PPO}); (2) demonstration-guided reinforcement learning with a proxy imitation objective~\cite{rajeswaran2018learning} (\emph{DAPG}); and (3) using adversarial objectives to keep the policy close to the demonstrations~\cite{peng2021amp} (\emph{AMP}). DAPG is the closest to our work, as it trains on a weighted sum of behavioral cloning and reinforcement learning losses. However, it assumes access to expert demonstrations in the downstream task. Our pre-training dataset does not fulfill this assumption. Indeed, humans might not behave optimally according to the reward, or the task might not be well represented in the pre-training dataset. Similarly to previous work~\cite{jiang2024transic}, we found that training from scratch is unsuccessful using joint-position control as action space, consistently leading the PPO baseline to fail. Therefore, we use the moving-average action space proposed by Petrenko et al.~\cite{petrenko2023dexpbt} to improve its performance. \revision{All baselines use the same environment settings and training strategy, \eg, domain randomization parameters, as our approach.}

\mypara{Tasks and Metrics.} We evaluate approaches on three tasks. The first requires picking objects and placing them at a specific location (\emph{Grasp and Lift}). The second is to grasp objects and throw them in a basket (\emph{Grasp and Throw}). In the final task, the robot is required to open a cabinet.
 We evaluate performance using success over 256 environments with different objects and report the mean and standard deviation over \secrevision{three} seeds per approach. More details can be found in the appendix.

% We found that PPO with our action space (joint position control) fails to learn. Indeed, joint position control is known to be unsuitable for exploration ~\cite{jiang2024transic} in RL. For a fair comparison, we use the action space for PPO as defined in \cite{petrenko2023dexpbt}. \textbf{DAPG}\cite{rajeswaran2018learning} computes gradients from expert demonstrations in addition to the policy gradient for each update. This requires access to the pre-training corpus during finetuning and also assumes task-aligned expert behavior in the demos. \textbf{AMP}\cite{peng2021amp} provides an additional reward to the agent to mimic the behavior in the demonstrations. This reward is computed using a discriminator learnt on the fly to differentiate between the trajectory rollouts from the agent and those from the pre-training corpus. An adversarial training objective makes it hard to train.

\begin{figure}[t]
    \centering
    \includegraphics[width=\textwidth]{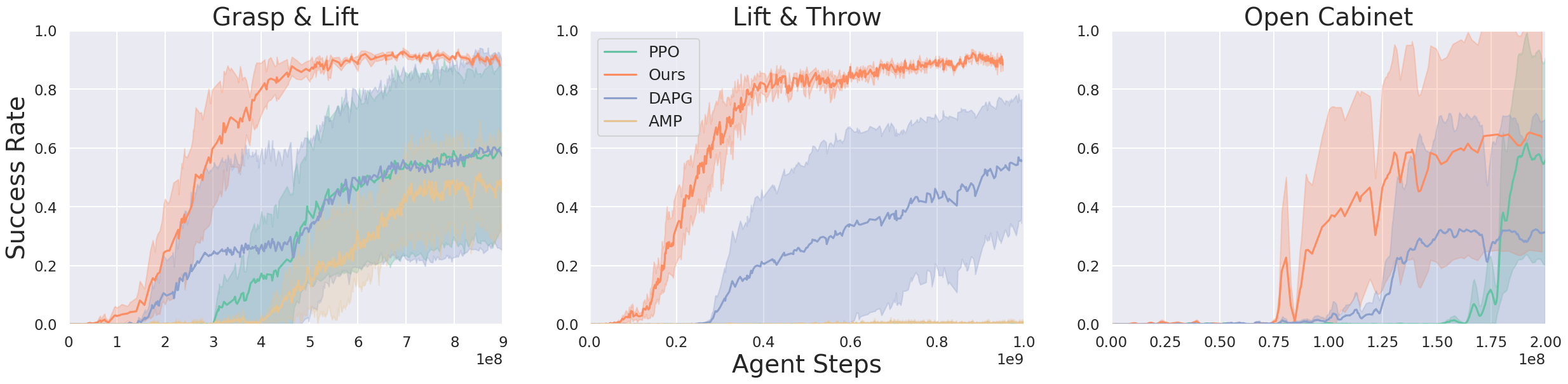}
    \caption{\footnotesize{\textbf{Comparison of HOP-initialized actor with baselines.} HOP improves sample-efficiency of online RL across multiple tasks, particularly when the downstream task and the behaviors in the data are less aligned, as in \textit{Lift \& Throw}. Runs are averaged across \secrevision{three} randomly chosen seeds.}}
    \label{fig:all_tasks}
\end{figure}

\mypara{HOP enables sample-efficient RL and effective exploration}
In Figure~\ref{fig:all_tasks}, it is demonstrated that our approach outperforms all baselines by a large margin, especially when the pretraining corpus is not closely related to the task. This is expected because DAPG strongly biases exploration in the neighborhood of the pre-training trajectories, which may potentially be misaligned with the downstream task. Furthermore, we observed that the adversarial training scheme of AMP is unstable and does not scale well with the amount of data. Finally, we find that using HOP leads to a 2-5X improvement in sample efficiency compared to training from scratch. 
Overall, these experiments show that initializing with HOP enables more informed exploration than the baselines and reduces variance in policy gradients. 

\label{sec:result}
\begin{figure}[t]
    \includegraphics[width=\textwidth]{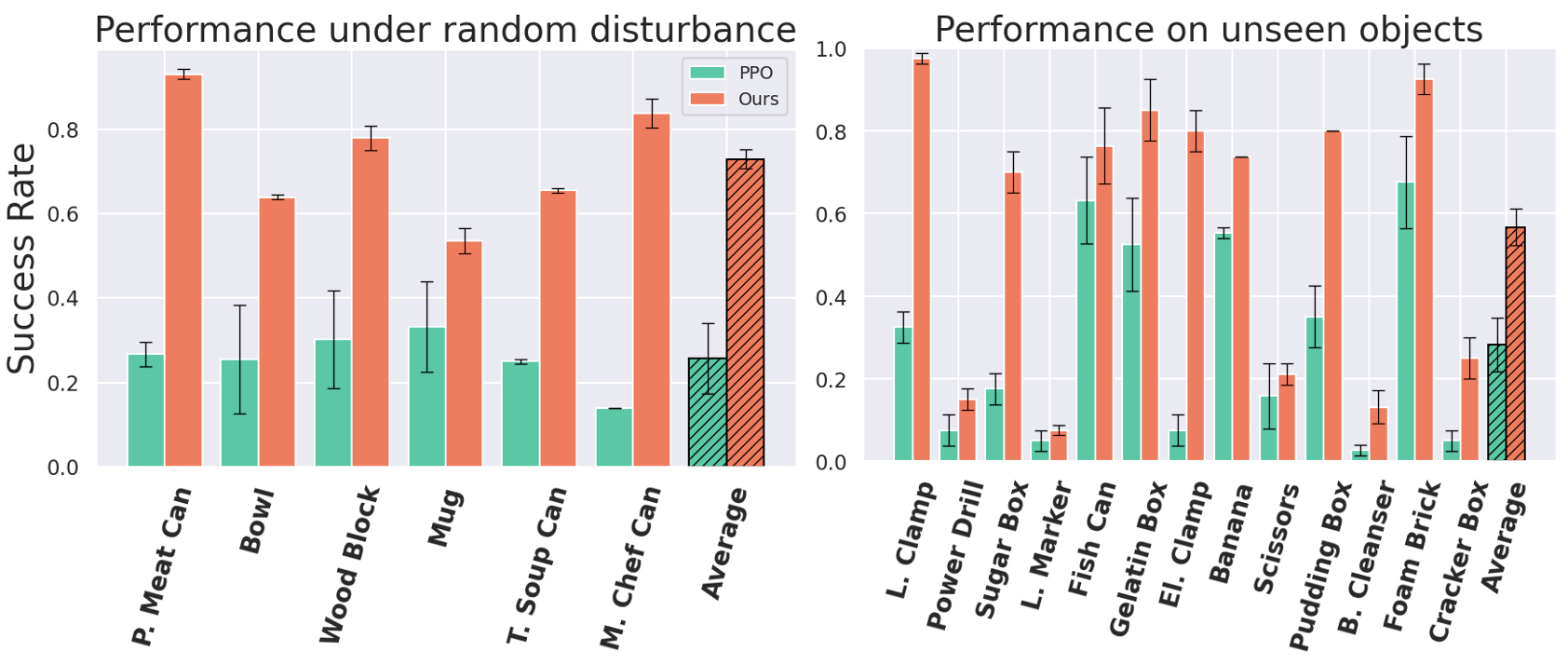}
    \caption{\footnotesize{\textbf{Evaluating RL finetuning under out-of-distribution scenarios} (Left)  To test grasp robustness in the task \textit{Grasp \& Lift}, we apply to the grasped objects, forces in random direction equal to their weights. When initialized with HOP, the resulting policy is more than $3\times$ more robust compared to training PPO from scratch. (Right) We evaluate grasp success on multiple objects from the YCB dataset that were not part of the training set. When initialized with HOP, the resulting policy is more than $2\times$ more robust compared to training PPO from scratch. }}
    \label{fig:ood}
\end{figure}
\begin{figure}[t]
    \centering
    \includegraphics[width=\textwidth]{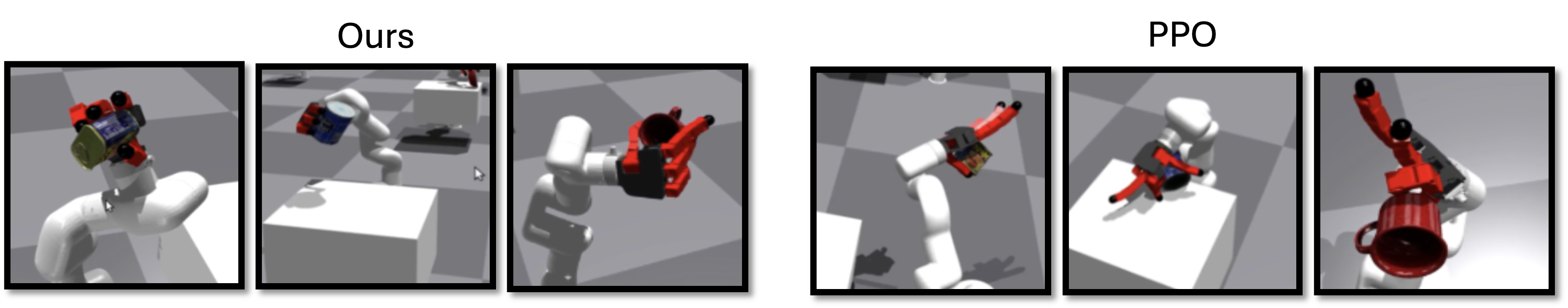}
    \caption{\footnotesize{Online exploration around the learnt prior from humans leads to grasps with more human-like and stable affordances compared to training PPO from scratch.}}
    \label{fig:affordance}
\end{figure}

\mypara{HOP learns robust and general behaviors}
% Behaviors learned by an embodied RL agent depend heavily on the extent of reward engineering. For a simplistic reward function, it tends to learn unnatural behaviors that are sensitive to disturbances \cite{rajeswaran2018learning}. 
Policies fine-tuned from HOP can potentially bias exploration toward human-like behavior, leading to more robustness against forces. This is shown in Fig.~\ref{fig:ood}. Agents trained with our approach perform better when subject to forces than the ones trained from scratch. In addition, we show in Fig.~\ref{fig:ood} that our approach generalizes 3x better than the policy trained from scratch. \revision{The training objects are different from the testing ones in their mass, aspect ratio, and relative size with respect to the hand. The performance generally drops whenever the test object is heavy (power drill), too large (cracker box), or too small (marker and scissors) for the allegro hand, which is approximately 1.5X larger than a human hand.} Note that in these experiments we train the scratch policy with two billion samples.

\mypara{Affordances} Training an RL policy from scratch for a dexterous hand often leads to grasping poses that are unlike general human affordances. Online RL exploration near a learned human-object interaction prior biases the optimization landscape to favor human-like affordances. As shown in Figure ~\ref{fig:affordance}, we find that for the \textit{Grasp and Lift} task, our policy grasps objects with more stable and human-like affordances than PPO training from scratch.

\subsection{Comparison to learning a hand-only motion prior (simulation)}
\label{sec:hand-only}

Prior work has shown the benefits of learning a prior on hand motions from videos of human activities~\cite{bharadhwaj2023zero,ze2024h,shaw2023videodex}. This section aims to understand the benefits of learning a prior on the object and the hand \emph{jointly}. We hypothesize that learning from hand-object interactions gives the base model useful information beyond eigen-grasps (which are captured by a hand-only motion prior), like, for instance, pre- and post-contact trajectories, intuitive physics of the interaction, and human preferences.

\begin{figure}[t]
    \centering
    \includegraphics[width=\textwidth]{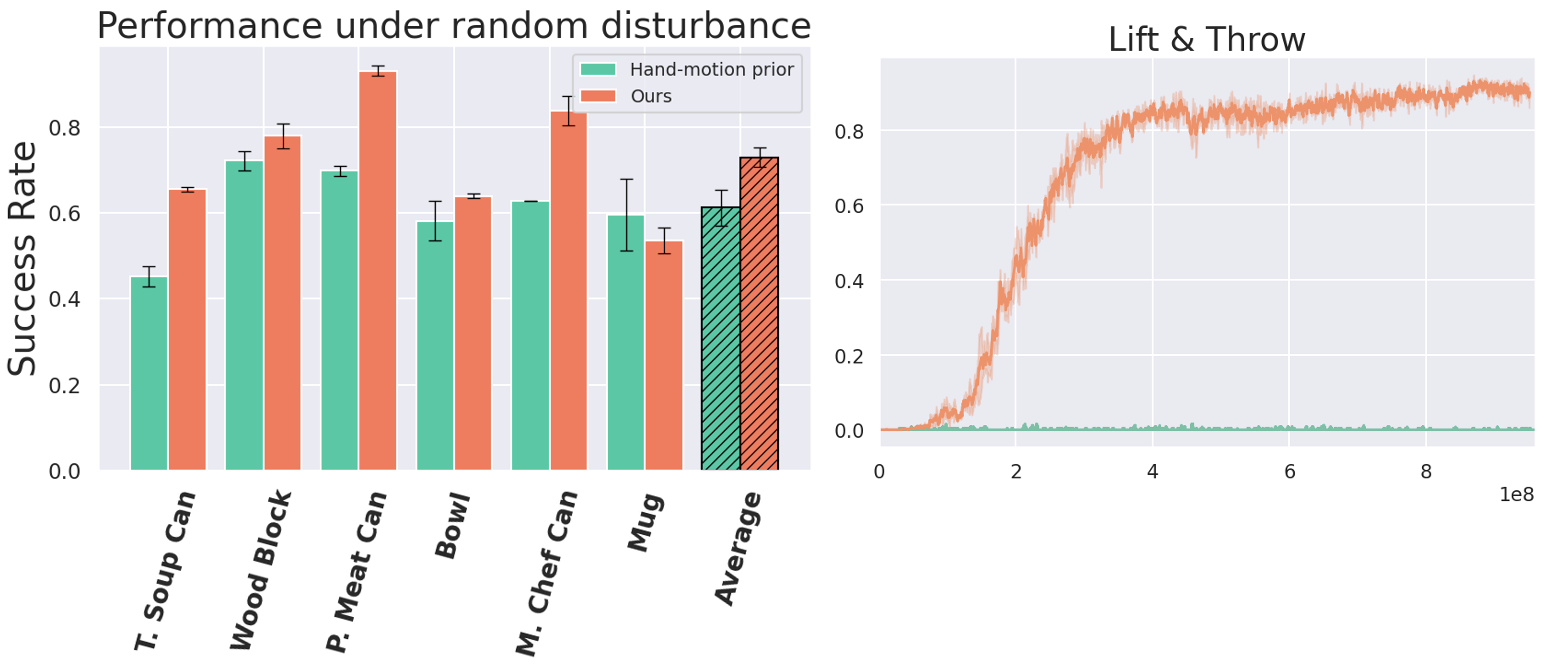}
    \caption{\footnotesize{Pre-training only a hand-motion prior leads to decrease in robustness of grasps to force disturbances (left). With our approach, the pre-trained policy learns a prior on object affordances which leads to more robust grasps. 
    \revision{In addition, pretraining with object poses leads to a more flexible prior and better finetuning to tasks less aligned with the pre-training data (right).}}}
    \label{fig:noobject}
    % \includegraphics[width=0.5\textwidth]{figures/throw_noobj.png}
    % \caption{\footnotesize{Pre-training only a hand-motion prior leads to decrease in robustness of grasps to force disturbances. With our approach, the pre-trained policy learns a prior on object affordances which leads to more robust grasps.}}
    % \label{fig:noobject}
    
\end{figure}

We evaluate this hypothesis by training a base policy on our pre-training corpus using masked object observations. This encourages the base policy to primarily learn a \textit{hand motion} prior. As illustrated in Figure \ref{fig:noobject} (left), we observe that such a pre-trained policy exhibits reduced robustness to grasp disturbances. Furthermore, we find that the hand-only prior is insufficient for learning an effective policy in the Grasp and Throw task (Fig.~\ref{fig:noobject}, right). Since this task is underrepresented in the pre-training corpus, the hand motions required are unlikely to be adequately captured by a hand-only prior. In contrast, learning a joint hand-object prior provides the model with a more comprehensive understanding of manipulation, enabling quicker adaptation to this downstream task.

\section{Related Work}

\mypara{Learning Policies from Human Videos.} In-the-wild videos hold the promise of solving the data problem in robotics. One of the pioneering efforts in this direction is by Yang et al.~\cite{yang2015robot}, where video data was used to generate action plans. Several works followed up on this idea, relying on pre-defined action primitives~\cite{nguyen2018translating, lee2017learning, arapi2018deepdynamichand, smith2019avid, bahl2023affordances,bharadhwaj2024track2act}. However, waving the requirement for pre-defined primitives is challenging since in-the-wild videos lack motor information. One way to recover motor information from videos is training with a trajectory-matching objective~\cite{bahl2022human,finn2016unsupervised,yu2018one,jin2019robot}, possibly using intermediate representations like object segmentation or optical flow~\cite{bharadhwaj2023towards,Wen2023AnypointTM,kumar2023graph}. However, this approach requires collecting task- and environment-specific videos where humans and robots operate in the same workspace. Therefore, the trajectory-matching formulation largely constrains the number of videos that can be used for training. To overcome these constraints, researchers have focused on either learning exclusively visual representations from videos or extracting 3D human poses and mapping them to robot actions. In the following, we cover these works in detail.

\mypara{Visual Representation Learning for Robotics.} Inspired by successes in computer vision~\cite{he2022masked} and natural language processing~\cite{devlin2018bert}, the robot learning community has recently focused on pretraining representations on large video datasets like Ego4D~\cite{grauman2022ego4d} and fine-tuning these representations on downstream tasks~\cite{Radosavovic2022, nair2022r3m, ma2022vip, majumdar2024we}. However, being the training objective based exclusively on image reconstruction, the representations focus primarily on low-level vision features, \eg, shapes or edges. This gives them limited benefits compared to representations trained on standard vision datasets~\cite{dasari2023unbiased}.
% Another approach to using visual representations from videos is finetuning a vision-language model with robot data~\cite{brohan2023rt}. 
Overall, these works focus on visual generalization, \eg, picking up two objects with the same shape but different colors. However, they have not yet demonstrated action generalization, where motor skills are adapted to accomplish novel objectives.
% Robot learning from human videos has been studied extensively in prior works. \red{mvp, r3m , vc-1} learn visual encoders using self-supervised learning on video datasets. These encoders are then used as fixed visual backbones for finetuning on downstream tasks. Our hypothesis is that augmenting videos with 3-D priors can allow us to learn more sensorimotor pre-trained representations than self-supervised learning on pixels. \red{This is empirically demonstrated by our experiments.}   

\mypara{Actions from Videos via 3D.} One common approach to extracting action information from videos is using 3D as an intermediate representation. If the embodiment gap is small, human motions can be mapped to robot actions via inverse kinematics. This is particularly effective when the videos are task-specific, \ie, when the robot aims to mimic the human motion~\cite{peng2018sfv,qin2022dexmv,ye2023learning,yang2022oakink,chen2022learning,chen2022dextransfer,patel2022learning}. Instead of learning specific skills, other works focus on learning a re-usable sensorimotor prior from videos. However, this prior only captures human actions~\cite{bharadhwaj2023zero,ze2024h,shaw2023videodex}, disregarding the trajectory of the manipulated object. Conversely, our work aims to use 3D hand-object interactions from in-the-wild videos to learn a re-usable prior for object manipulation.

\textbf{Dexterous Manipulation.} Dexterous manipulation has been studied for decades~\cite{fearing1986implementing,han1998dextrous,okamura2000overview,ciocarlie2009hand,morgan2022complex}. In recently years, learning-based approaches make significant progress~\cite{openai2018learning,yu2022dexterous}. They can be generally categorized to learning in simulation and then transferring to the real world (Sim-to-Real)~\cite{openai2019solving,handa2023dextreme,chen2023visual,qi2022hand,qi2023general,agarwal2023dexterous}, and learning in the real-world~\cite{qin2022from,wang2024dexcap,guzey2023dexterity,guzey2023see,arunachalam2023dexterous}. ~\citet{qin2022dexmv} uses hand-object trajectories but only use it to collect demonstrations. ~\citet{xu2023unidexgrasp} also does functional grasp generation, but the results are limited in simulation. Most of the aforementioned work learn policies from scratch and does not use any internet data as prior. Our approach studies learning hand-object interaction prior from human videos and is effective both in simulation and real-world.

\section{Conclusion and Limitations}
\label{sec:conclusion}
This work presents an approach to learning general yet flexible manipulation priors for robot policies from human videos. While our approach demonstrates a way to pre-train on a single object interaction, this can, in practice, be limiting. Indeed, human behavior in a video can potentially be conditioned on information encompassing multiple objects in the current and previous scenes. This leads to a loss of signal that could be extracted from the raw video. We predict that advances in 3-D reconstruction will enable us to use a more complex scene reconstruction and pretraining.
	
%===============================================================================
%===============================================================================

% The acknowledgments are automatically included only in the final and preprint versions of the paper.
\acknowledgments{This work was supported by the DARPA Machine Common Sense program, the DARPA Transfer from Imprecise and Abstract Models to Autonomous Technologies (TIAMAT) program, and by the ONR MURI award N00014-21-1-2801. This work was also funded by ONR MURI N00014-22-1-2773. We thank Adhithya Iyer for assistance with teleoperation systems, Phillip Wu for setting-up the real robot, and Raven Huang, Jathushan Rajasegaran and Yutong Bai for helpful discussions.}

%===============================================================================

\bibliography{example}  % .bib

\newpage

\section{Supplementary Material}

\subsection{3D Hand-Object Interaction from Videos}

Our setup closely follows the pre-trained MCC-HO model~\cite{wu2024reconstructing} to lift RGB videos to 3D. 
However, this approach was designed to work on a single frame, while we are interested in extracting trajectories from videos.
Specifically, MCC-HO input is the image patch containing the hand and manipulated object.
Acquiring these patches requires identifying which hand and object is part of the manipulation sequence from in-the-wild videos.
To make the problem tractable, we use the fact that the 100 Days of Hands dataset (100DOH)~\cite{shan2020understanding} includes 100K labeled frames (\eg, hand and object bounding boxes) randomly sampled from Internet videos.

We use such sparse labels as follows:
Given a video with a labeled frame at timestamp $t_0$, we download a 10-second video clip centered at $t_0$ using the original video frame rate. Then, we propagate the combined hand-object bounding box from $t_0$ to $t < t_0$ and $t > t_0$ iteratively until the interacting hand is no longer detected by HaMeR~\cite{pavlakos2024reconstructing}.
At each frame $t$, the labeled hand-object bounding box is translated so the hand bounding-box center is aligned with the HaMeR hand bounding-box center.
Subsequently, the images are cropped/resized using these hand-object bounding boxes and passed to MCC-HO for network inference.
This gives a sequence of 3D hand-object poses.
We do not temporally smooth the sequences further (\eg, via high-pass filtering) since this happens (to a certain extent) as a by-product of the robot's inertia during the simulator-in-the-loop motion re-targeting.

\subsection{Simulation Experiments Setup}

We reuse environment definitions from previous works (~\cite{makoviychuk2021isaac, xiao2022masked}) with minimal changes.  We do not make any changes to the structure of the reward function. Below, we provide a brief description for each task: 
\begin{enumerate}
    \item \textit{Grasp and Lift} We adapt \href{https://github.com/isaac-sim/IsaacGymEnvs/blob/main/isaacgymenvs/tasks/allegro_kuka/allegro_kuka_regrasping.py}{this} environment definition from IsaacGymEnvs\cite{makoviychuk2021isaac} for this task. We changed the robot to an Xarm7 with an Allegro Hand as the end-effector attached vertically. For increasing realism, we enable gravity in the environment and increase the number of convex decompositions in VHACD. For the objects, we use 6 canonical objects from the YCB\cite{calli2017yale} benchmark. Specifically, we use the \textit{Tomato Soup Can}, \textit{Wood Block}, \textit{Potted Meat Can}, \textit{Bowl}, \textit{Bowl}, \textit{Master Chef Can} and \textit{Mug}. The choice of the objects is done keeping in mind the limitations of the robot embodiment such as the large size of the palm and thick fingers.
    \item \textit{Lift and Throw} We adapt \href{https://github.com/isaac-sim/IsaacGymEnvs/blob/main/isaacgymenvs/tasks/allegro_kuka/allegro_kuka_throw.py}{this} environment definition for \textit{Lift and Throw}. We change the robot, the objects and the physical parameters same as in \textit{Grasp and Lift}.
    \item \textit{Open Cabinet} We adapt \href{https://github.com/ir413/mvp/blob/master/pixmc/tasks/kuka_cabinet.py}{this} environment definition from PixMC\cite{xiao2022masked} for \textit{Open Cabinet}. We change the robot as in above tasks and keep other parameters the same. The input to the policy in this task is the point cloud for the handle of the cabinet.
\end{enumerate}

\subsection{Model Architecture}
Our policy is a GPT-2 style transformer with causal attention. The transformer has 4 heads, 4 layers, and a hidden dimension of 192. For real-world experiments, proprioception and the depth images are embedded to the hidden size using a linear projection and a 4-layer CNN, respectively. In simulation, point clouds are embedded to the hidden size using a pointnet with 2 hidden layers of size 64.
The input to the transformer are proprioception and depth/pointcloud embeddings for previous 16 timesteps. Additive learnable positional embeddings are used for both the proprioception and depth embeddings.

\subsection{Real-World Experiments Setup}

\mypara{Data Collection} We build a custom teleoperation setup for data collection by combining two existing systems. We control the xArm with a Gello~\cite{wu2023gello} and the hand motion with VR using OpenTeach~\cite{iyer2024open}. We experimented with controlling the whole system via VR but found obtaining precise and smooth motion challenging. While our solution could achieve such motions, it comes with the disadvantage of requiring two people to collect demonstrations. \revision{We randomize the initial pose of the robot between demonstrations by adding a uniform noise of magnitude 0.3 rad to all joints from a fixed starting location. }

\mypara{Inference} While sensors operate at different frequencies, we get the latest available measurements from each sensor at constant intervals to achieve a whole inference loop of 20Hz. The predicted action, \ie absolute joint positions, are given to a low-level P controller that operates at 120Hz. Such controller directly sends commands to the xArm API to convert joint position into joint torques.

\mypara{Tasks} We study three tasks of increasing complexity. We recommend the reader to checkout videos on our supplementary website to visualize the task setup. For all tasks, we report a success if the policy can complete the task in less than 30 seconds. In the following, an explanation of each task:
\begin{itemize}
    \item \emph{Grasp and Drop} The robot needs to pick up a soft block and drop it in a bowl. The box and the bowl are both at the same location at training and evaluation. Note that for this task, playing a demonstration open loop has almost 100\% success rate.
    \item \emph{Grasp and Pour} The robot needs to pick up a bottle and rotate it to point its cap towards the bowl. There are two bottles of similar shape but very different material, one 3D printed and another of plastic (filled with water), that we use for data collection and evaluation. The bowl and bottle position are constant at training and evaluation time. For this task, replaying a demonstration open-loop is successful, but more sensitive to the bottle location. Tiny variations in position will make open-loop fail. However, the trained policies are robust to such small variations.
    \item \emph{Grasp and Lift} In this task, the robot should grasp one object and lifting it a minimum of 15 cm above the table. There are four very different objects we work on, with different material and shape. In this task, replaying a trajectory does not work since the object to be grasped is not known in advanced. The objects and their location are same at training and test time. 
\end{itemize}

\mypara{Training Details}
\label{sec:training_details}
\begin{itemize}
    \item \mypara{Ours} We finetune the model with a batch size of 128 using the AdamW optimizer with learning rate set to 1e-4 and weight decay set to 1e-2. For each task, we finetune for 9000 gradient steps and pick the best checkpoint from those collected every 1000 steps. For comparing with our base model trained from scratch, we train the policies trained from scratch with the same configuration as above. 

    \item \mypara{Diffusion Policy} For training the diffusion policy baseline, we keep the batch size, optimizer and other parameters same as in \cite{chi2023diffusion}. We train diffusion policy for 60,000 gradient steps for each task. We find the policy trained for 15,000 gradients steps to be performing similar to later checkpoints for all tasks.

    \item \mypara{Imagenet Baselines} For the imagenet baselines, we train for 40,000 gradient steps and pick the best checkpoint every 5,000 step with the same optimizer as ours.

    \revision{
    \item \mypara{Visual representation learning baselines} We compare with R3M, MVP and VIP from this line of approaches. For each baseline, we replace the image encoder in our architecture with a frozen instance of one of these visual encoders. For training with RGB images, we apply data augmentation in the same way as in \cite{chi2024universal}. We train each model for 40000 gradient steps and choose the best checkpoint every 5,000 steps.  
    }
     
\end{itemize}

\end{document}